\documentclass[10pt,twocolumn,letterpaper]{article}

\usepackage{cvpr}
\usepackage{times}
\usepackage{epsfig}
\usepackage{graphicx}
\usepackage{amsmath}
\usepackage{amssymb}
\usepackage{bm}
\usepackage{multirow}
\usepackage{multicol}
\usepackage{xcolor}
\newcommand\todo[1]{\textcolor{black}{#1}} 

\begin{document}

\title{A Large-Scale Re-identification Analysis in Sporting Scenarios: \\the Betrayal of Reaching a Critical Point}

\author{David Freire-Obregón, Javier Lorenzo-Navarro, Oliverio J. Santana, \\Daniel Hernández-Sosa, Modesto Castrillón-Santana\\
Universidad de Las Palmas de Gran Canaria\\
{\tt\small david.freire@ulpgc.es}
}

\maketitle
\thispagestyle{empty}

\begin{abstract}

Re-identifying participants in ultra-distance running competitions can be daunting due to the extensive distances and constantly changing terrain. To overcome these challenges, computer vision techniques have been developed to analyze runners' faces, numbers on their bibs, and clothing. However, our study presents a novel gait-based approach for runners' re-identification (re-ID) by leveraging various pre-trained human action recognition (HAR) models and loss functions. Our results show that this approach provides promising results for re-identifying runners in ultra-distance competitions. Furthermore, we investigate the significance of distinct human body movements when athletes are approaching their endurance limits and their potential impact on re-ID accuracy. \todo{Our study examines how the recognition of a runner's gait is affected by a competition's critical point (CP), defined as a moment of severe fatigue and the point where the finish line comes into view, just a few kilometers away from this location. We aim to determine how this CP can improve the accuracy of athlete re-ID. Our experimental results demonstrate that gait recognition can be significantly enhanced (up to a 9\% increase in mAP) as athletes approach this point. This highlights the potential of utilizing gait recognition in real-world scenarios, such as ultra-distance competitions or long-duration surveillance tasks.
}
\end{abstract}

\section{Introduction}

Gait refers to the distinctive manner in which an individual walks or runs and is a complex biometric that involves space and time. Although gait is not considered highly distinctive, it can still be sufficiently discriminatory to enable verification in specific low-security applications. As a behavioral biometric, gait may not remain consistent over extended periods, mainly due to fluctuations in body weight, major joint or brain injuries, or intoxication.

In the pioneering work conducted by Jain et al.~\cite{Jain04}, it was reported that collecting gait data is comparable to capturing facial images and can serve as a viable biometric modality. The authors noted that gait-based systems need extensive video footage of a person walking to capture multiple movements in each articulate joint, resulting in a high input requirement and computational cost. However, appearance-based HAR models have shown remarkable performance at a reasonable computational cost in recent years.

\begin{figure}[bt] 
\begin{minipage}{1\linewidth}
    \centering
    \includegraphics[scale=0.5]{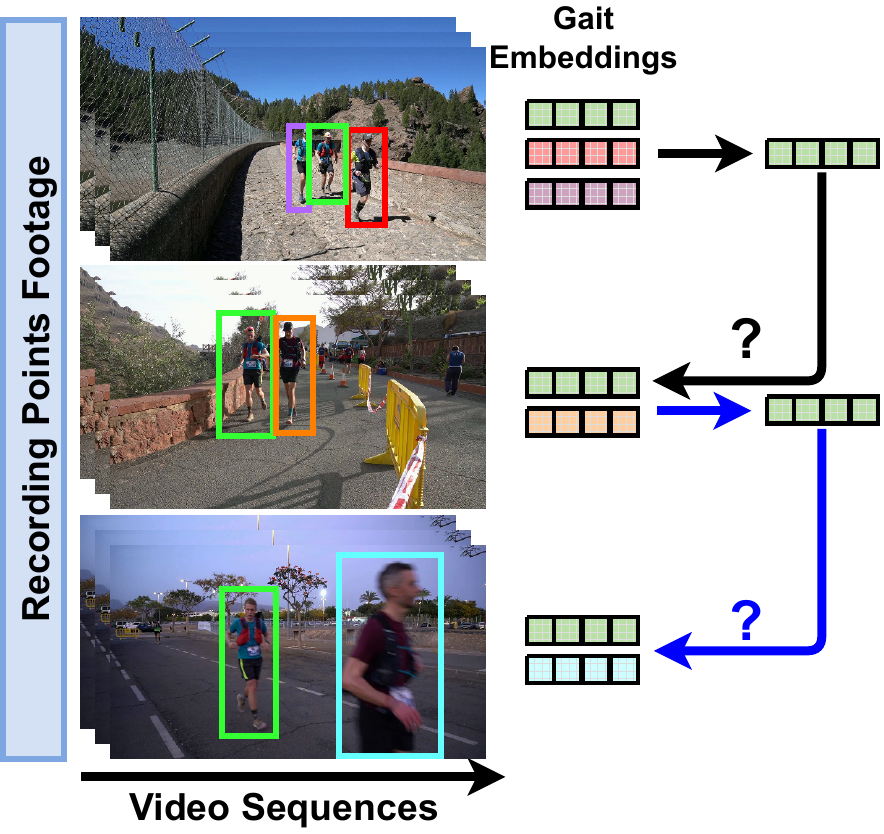}
    \caption{\textbf{Analyzing re-ID during the race course.} We collect footage samples of a runner at each recording point, e.g., the runner enclosed in a green container in the image. Utilizing various pre-trained encoders (backbones), we extract embeddings of the runner from the footage.  These embeddings are then inputted into a model predicting the runner's identity in the next recorded-point video set. \label{fig:problem}}
    \end{minipage}
\end{figure}

In a sporting context, athlete re-ID is a crucial component, as it enables the tracking and monitoring of athletes throughout a competition or event~\cite{penate20prl}. In addition to enhancing security and safety, athlete re-ID can provide valuable information for coaches, fans, and media outlets.

One key advantage of athlete re-ID is the ability to track athlete performance throughout a competition. Using computer vision algorithms to identify individual athletes across multiple cameras, coaches and trainers can monitor their athletes' form, technique, and progress~\cite{Cioppa22}. This information can be used to optimize training regimes, improve performance, and prevent injury. In addition to enhancing athletic performance, athlete re-ID can improve the fan experience. By identifying individual athletes, fans can receive real-time updates on their favorite players, including statistics, highlights, and interviews~\cite{Manafifard17}. This information can be delivered through various channels, including mobile apps, social media, and live broadcasts. Furthermore, athlete re-ID can enable personalized services for fans, such as seat upgrades, merchandise discounts, and VIP access.

Another advantage of re-ID in this scenario is the ability to provide valuable insights for media outlets. By identifying individual athletes across multiple events, media outlets can track trends and patterns in performance and identify emerging talent~\cite{Torres22}. This information can be used to create compelling stories and analyses for sports fans and inform betting and fantasy sports decisions. Despite its potential benefits, re-ID in sports tackles several challenges, including technical limitations and ethical considerations~\cite{Enck10}. For example, athlete re-ID systems may need help differentiating between athletes with similar physical attributes or in situations with large crowds or low lighting. 

Additionally, human gait is considered a distinctive and hard-to-replicate characteristic, in other words, a crucial biometric feature for identification purposes~\cite{Nambiar19}. Therefore, gait recognition has been identified as an essential technology for diverse applications in high-security environments, as well as public areas like airports, stations, and banks~\cite{Hossain12}. In this study, we aim to advance the field by conducting runner re-ID in a real-world ultra-distance competition setting. We seek to address the following inquiries: To what extent can the embeddings of appearance-based HAR models be applied to a sports re-ID context? Can gait offer valuable re-ID perspectives? How does physical exhaustion impact athletes' gait for re-ID objectives? In this regard, we have analyzed eleven HAR pre-trained models (backbones) in a dataset collected to evaluate runner re-ID methods in real-world scenarios (see Figure~\ref{fig:problem}). The obtained outcomes are significant (up to 63.3\% of mAP) and have led to insightful observations. The models based on HAR, which generate embeddings from a reduced number of frames (such as Slow channel and I3D models), outperform the models that do not. Furthermore, identifying athletes is enhanced by 6\% to 9\% when considering the impact of \todo{reaching the CP}.

\section{Related Work}

Compared to other biometric traits, gait possesses unique advantages. Its most notable feature is its non-intrusiveness, meaning it can be captured from a distance without the subject's consent, unlike other biometrics such as fingerprints, face, hand geometry, iris, voice, and signature, which require physical contact or proximity to the recording probe 
\cite{Galbally14, Jain04, Jiang16}. Additionally, gait is more challenging to conceal, steal, or forge. It may change over time due to factors such as fluctuations in body weight, or major injuries to joints or the brain. Hence, gait recognition holds tremendous promise in the domains of crime investigation, social security, and access control. As a visual identification task, the primary objective of gait recognition is to acquire distinct and unchanging features from the constantly changing attributes of the human body shape over time. Nonetheless, in real-world scenarios, variations like carrying bags, wearing coats, and camera viewpoints can cause substantial changes in gait appearance, making gait recognition a challenging task~\cite{Fan20}. \todo{Furthermore, our analysis suggests that reaching the CP can potentially lead to distortions in human gait, which may impact the performance of re-ID systems. As a result, we aim to investigate how CP affects human gait and its implications for re-ID.}

\begin{figure*}[t]  
\begin{minipage}[t]{1\linewidth}
    \centering
    \includegraphics[scale=0.7]{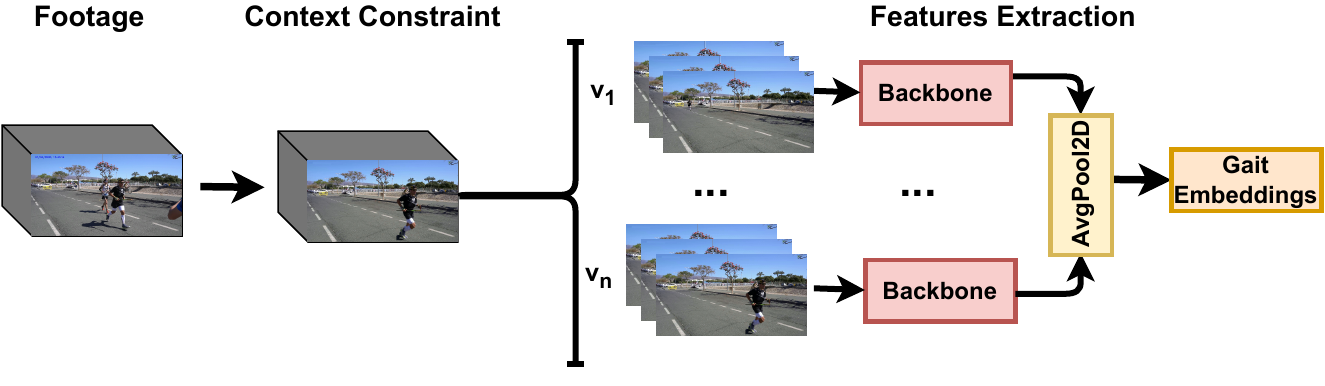}
      \end{minipage}
    \caption{\textbf{Gait features extraction overview.} The process comprises two main components: the footage pre-processing and regression blocks. Initially, the tracker assists in isolating the runner's activity from the background in the first stage. Next, the footage is divided into~$n$ small clips through down-sampling, and then a pre-trained human-action model processes these clips to extract features. These features are synthesized using an average pooling method, producing a final tensor for input to the classifier.
     \label{fig:pipeline}}
\end{figure*}

According to Teepe et al., two main categories of spatial feature extraction methods exist: appearance-based and model-based~\cite{Teepe21}. Appearance-based methods extract a binary human silhouette image from the original image, usually obtained through background subtraction for static scenes but more challenging for dynamic settings. Approaches using the whole shape as input are prevalent, but current methods focus on specific body parts. On the other hand, model-based methods consider the body's physical structure, extracting features from model data, often containing handcrafted velocity, angle, and other measurements~\cite{Liu19,Wang03}. Similarly, the classification of HAR models can be categorized based on the representation of the human pose, which is crucial for evaluating the performed action. In their work, Lei et al. have identified three primary pose representations for action quality assessment~\cite{Lei19}. However, the challenge lies in identifying robust features from pose sequences and establishing a method to measure pose feature similarity~\cite{Wnuk10}. On the other hand, skeleton-based representations encode the relationship between joints and body parts~\cite{Freire20}. Nevertheless, the estimated skeleton data can be noisy in realistic scenarios, mainly due to occlusions or changing lighting conditions~\cite{Carissimi18}, such as those encountered in an ultra-distance race under wild conditions.

Ultra-distance races use Radio Frequency Identification (RFID) tags worn by runners on their shoes, wristbands, or bibs to track their progress with readers placed throughout the course~\cite{Chokchai18}. Despite these measures, some runners still cheat by taking shortcuts~\cite{Zakrzewski23}. Therefore, re-ID is a primary concern in the research surrounding this sport. Mathematical models have been proposed~\cite{Goodrich21}, as well as invasive RFID-like devices~\cite{Lingjia19}, and non-invasive models based on the runner's appearance~\cite{Klontz13}, bib number~\cite{Ben-Ami12-bmvc,Carrascosa20}, or their arm-swing pattern \cite{Yapkan21}.

This study advances the field of athlete re-ID by incorporating HAR models on sequences of runners. Lately, the analysis of body movements has also been utilized to determine additional human indicators, specifically for recognizing emotions. Ahmed et al. presented a two-layer feature selection framework for emotion classification that accurately recognizes five basic emotions (happiness, sadness, fear, anger, and neutral) using a comprehensive list of body movement features~\cite{Ahmed20}.

\section{Method}

This section comprises three main subsections: pre-processing, pre-trained encoders (backbones), and classifier and loss functions. In the pre-processing step, we apply a context constraint to improve focus on the athlete in the input video frames. This helps to remove irrelevant background information and reduce noise in the feature representation. The pre-trained encoders (backbones) extract gait embeddings from the pre-processed frames. This subsection describes the pre-trained HAR models considered in our experiments, including their architecture and input requirements.
The classifier and losses subsection describes the re-ID model used to match athlete identities across different video frames. We use a triplet loss, which aims to minimize the distance between embeddings of the same identity and maximize the distance between embeddings of different identities. This loss is known to be effective in training re-ID models with appearance-based features. We also consider a quadruplet loss, which extends the triplet loss by considering the hardest negative example.

\subsection{Pre-processing}
\label{sec:preprocessing}
We have developed a modular multi-stage pipeline to extract gait embeddings, as shown in Figure~\ref{fig:pipeline}.
To improve the quality of the embeddings, it is necessary to provide clean footage input to the action recognition networks~\cite{freire22icpr}. This means removing elements such as other athletes, race staff, and moving vehicles from the scene, which are irrelevant to the gait analysis. Therefore, the initial block pre-processes the raw input data to focus on the runner of interest. We used ByteTrack~\cite{zhang2021bytetrack}, a multi-object tracking network, to track the runner in each footage. Then, we applied context-constrained pre-processing to generate the scenario for our experiments.
To obtain the context-constrained footage for a runner $i$ at a given time $t \in [0,T]$ and in a recording point, $RP \in [0,P]$, we used the runner's bounding box area $BB_i(t, RP)$ and the average number of frames $\tau(RP)$ needed to generate clean footage where the runner appears with a still background. The new pre-processed footage $F'_i[RP]$ is obtained as follows: 

\begin{equation} 
\label{eq:contextremoval} 
F'_i[RP]=BB_{i}(t, RP) \cup \tau(RP) 
\end{equation}

As depicted in Figure~\ref{fig:pipeline}, the given input footage, consisting of \textit{n} frames, undergoes downsampling and is divided into \textit{n} video clips (\textit{{$v_1$, ..., $v_n$}}), each containing $q$ consecutive frames that capture an activity snapshot (refer to Figure~\ref{fig:pipeline}). These video clips are passed through a pre-trained encoder for HAR, resulting in a \mbox{p-dimensional} feature vector. These models have been pre-trained on the Kinetics dataset~\cite{Kay17}, which includes $400$ action categories. After obtaining the feature vectors for all \textit{n} video clips, an average pooling layer ensures that all clips contribute equally. Finally, the extracted features are fed into triplet, or quadruplet network, to obtain the embedding.

\subsection{Pre-trained encoders}
\label{sec:backbones}
In this section, we will provide a summary of the characteristics of the eleven human-action recognition backbones that were considered.

\textbf{C2D}. The Convolutional 2D (C2D) model processes 2D spatial images, each representing a video clip frame~\cite{C2D14}. It uses a convolutional neural network (CNN) architecture similar to image classification. Typically, the C2D model comprises several convolutional layers followed by pooling layers that extract increasingly complex features from the input frames. The convolutional layers apply learned filters to the input frames, producing feature maps that capture spatial information. The pooling layers then downsample the feature maps, reducing the spatial resolution while retaining the most salient features. The resulting feature maps are then flattened into a feature vector.

\textbf{I3D}. The Inflated 3D ConvNet (I3D) model operates on short clips of video frames represented as 3D spatiotemporal volumes~\cite{Carreira17}. It employs a two-stream approach, where one stream processes RGB images and the other processes optical flow images, capturing appearance and motion cues. The RGB stream is initialized with weights pre-trained on large-scale image classification datasets such as ImageNet, while the flow stream is randomly initialized and fine-tuned along with the RGB stream. 
The output of the last layer of the I3D model is a feature vector that summarizes the appearance and motion information of the input video clip.

\textbf{I3D NLN}. The Non-local Network (I3D NLN) is a modified version of the I3D model that integrates non-local operations for better video spatiotemporal dependency modeling~\cite{NonlocalNN17}. Like I3D, I3D NLN operates on 3D spatiotemporal volumes and adopts a two-stream architecture with RGB and optical flow streams. However, instead of the Inception module, I3D NLN uses non-local blocks that can learn long-range dependencies between any two positions in the input feature maps. The non-local block computes a weighted sum of input features from all positions based on the similarity between every other position in the feature maps. This allows I3D NLN to capture global context information, leading to improved modeling of temporal dynamics. 

\textbf{Slow} is a HAR model that uses a two-stream architecture to capture short-term and long-term temporal dynamics in videos~\cite{Slow21}. The slow pathway processes high-resolution frames at a lower frame rate. Slow is similar to the C2D model and includes a temporal-downsampling layer to capture longer-term temporal dynamics. 

\textbf{SlowFast} includes a slow pathway that processes high-resolution frames at a slower frame rate, capturing spatial information and long-term temporal structure~\cite{SlowFast19}. The fast pathway processes low-resolution frames at a faster frame rate, capturing fine-grained motion information and short-term temporal structure. The slow pathway involves a deep 3D CNN that processes each frame in a video sequence with a temporal stride of 16 frames. The fast pathway consists of a shallower 3D CNN that processes every other frame with a temporal stride of 2 frames. To produce the final video-level representation, the outputs of the two pathways are combined through a fusion module that uses a weighted sum of the features.

\textbf{X3D}. The Xception3D (X3D) is a 3D CNN that processes data in both spatial and temporal dimensions~\cite{X3D20}. It comprises a series of 3D convolutional blocks containing 3D convolutions, temporal and spatial convolutions, and nonlinear activations. These blocks are arranged hierarchically, with increased feature map size and decreased spatial resolution. X3D delivers four different model sizes: X3D-XS, X3D-S, X3D-M, and X3D-L. Each of these models has a different number of parameters and computational cost. The X3D-XS model is the smallest and fastest, whereas the X3D-L model is the largest.

\subsection{Loss functions}

The gait embeddings undergo processing by a small neural network that consists of two 512 dense layers separated by a batch normalization layer. The resulting new embeddings are designed to discriminate between identities and are forced to exist on a d-dimensional hypersphere via L2 normalization of the final output. The distance between gait embeddings is computed using the L2 distance function adopted by Schroff et al.~\cite{Schroff15}, which calculates the squared difference between feature vectors ($p$ and $q$) and sums them up. This distance measure is used to compute the loss function, which depends on the distances between embeddings of multiple input samples, depending on the chosen loss function. We have evaluated two loss functions: triplet loss and quadruplet loss.

Triplet loss compares three samples: an anchor sample, a positive sample (with the same identity as the anchor), and a negative sample (with a different identity from the anchor)~\cite{Schroff15}. The triplet loss function can be defined as follows:
\begin{equation}
\label{eq:TL}
\begin{aligned}
L_{triplet}(D_1, D_2) = & \max(D_1^2 - D_2^2 + m_1, 0) 
\end{aligned}
\end{equation}
where $D_1$ and $D_2$ are the distances $<anchor - positive>$ and $<anchor - negative>$, respectively. The margin parameter is denoted as $m_1$. The goal is to minimize the distance between the anchor and positive samples while maximizing the distance between the anchor and negative samples. 

Contrary to triplet loss, quadruplet loss compares four samples: an anchor, a positive, and two negatives~\cite{Chen2017}. The goal is to minimize the distance between the anchor and positive samples while maximizing the distance between the anchor and the two negative samples. This additional negative sample helps to increase the separation between different identities, leading to improved performance on tasks such as person re-ID. The quadruplet loss function can be defined as follows:

\begin{equation}
\label{eq:QL}
\begin{aligned}
L_{quadruplet}(D_1, D_2, D_3) = & \max(D_1^2 - D_2^2 + m_1, 0) \\
& + \max(D_1^2 - D_3^2 + m_2, 0) 
\end{aligned}
\end{equation}
where $D_1$, $D_2$, and $D_3$ correspond to the distances \mbox{$<anchor-positive>$}, \mbox{$<anchor-negative\,1>$}, and \mbox{$<negative\,1-negative\,2>$}, respectively. The margin parameters are denoted as $m_1$ and $m_2$. Accordingly, similar objects are closer together while dissimilar objects are pushed away from each other. Adding an additional negative sample distance ($D_3$) to the loss function in equation~\ref{eq:QL} can aid the network in learning a more generalized rule for similarity. Overall, the quadruplet loss is a more complex and computationally intensive approach than the triplet loss, but it can provide better performance in some cases.
In the presented experiments, we also explore using the quadruplet loss architecture to update the network parameters.

\begin{table*}[h!]
\caption{Mean average precision achieved by each considered backbone. The table is organized regarding backbones and losses when the latter is investigated. The second column shows the number of frames the backbone requires to make a prediction. Moreover, two competition stages are analyzed using scheme A$\rightarrow$B, where A stands as the RP considered for the probe and B as the gallery. The average columns show the mean mAP for each backbone, whereas the last row shows the mean mAP at each stage.}
\label{table1}
\centering
\begin{tabular}{||l|c||c|c|c||c|c|c||} 
 \hline
  Backbone & \#Frames  & \multicolumn{3}{|c||}{Triplet loss mAP} & \multicolumn{3}{|c||}{Quadruplet loss mAP}\\ 
 \cline{3-8}
   &  & RP1$\rightarrow$RP2$\uparrow$ & RP2$\rightarrow$RP3$\uparrow$ & Average$\uparrow$ & RP1$\rightarrow$RP2$\uparrow$ & RP2$\rightarrow$RP3$\uparrow$ & Average$\uparrow$ \\ 
 \hline\hline
  C2D~\cite{C2D14} & 8 & 45.9\% & 64.7\% & 55.3\% & 47.6\% & 56.0\% & 51.8\%\\
 \hline
 I3D~\cite{Carreira17} & 8 & 56.5\% & 68.2\% & 62.4\% & \textbf{56.5\%} & \textbf{67.2\%} & \textbf{61.9\%}\\
 \hline
 I3D NLN~\cite{NonlocalNN17} & 8 & 56.8\% & 62.0\% & 59.4\% & 52.6\% & 61.4\% & 57.0\%\\
 \hline
 Slow4x16~\cite{Slow21} & 4 & \textbf{60.1\%} & \textbf{66.6\%} & \textbf{63.3\%} & 54.6\% & 62.3\% & 58.5\%\\
 \hline
 Slow8x8~\cite{Slow21} & 8 & 55.7\% & 63.1\% & 59.4\% & 53.2\% & 61.6\% & 57.4\%\\
 \hline
 SlowFast4x16~\cite{SlowFast19} & 32 & 59.6\% & 62.2\% & 60.9\% & 53.0\% & 64.3\% & 58.6\%\\
 \hline
 SlowFast8x8~\cite{SlowFast19} & 32 & 54.2\% & 61.6\% & 57.9\% & 46.0\% & 59.6\% & 52.8\%\\
 \hline
 X3D XS~\cite{X3D20} & 4 & 51.5\% & 51.8\% & 51.6\% & 50.1\% & 55.0\% & 52.6\%\\
 \hline
 X3D S~\cite{X3D20} & 4 & 52.0\% & 51.9\% & 52.0\% & 49.9\% & 53.8\% & 51.9\%\\
 \hline
 X3D M~\cite{X3D20} & 13 & 49.5\% & 51.8\% & 51.8\% & 48.1\% & 57.8\% & 52.9\%\\
 \hline
 X3D L~\cite{X3D20} & 16 & 43.1\% & 51.6\% & 47.4\% & 43.8\% & 52.9\% & 48.3\%\\
 \hline
 \hline
 \multicolumn{2}{|c||}{\textit{RPs Average}} & 53.2\% & 59.6\% & - & 50.5\% & 59.3\% & -\\
 \hline
\end{tabular}
\end{table*}

\section{Experiments}

\subsection{Dataset}
In our study, we partially used the dataset introduced by Penate et al.~\cite{penate20prl}. This dataset was collected during the Transgrancanaria (TGC) 2020 ultra-distance running competition, which included up to six different distances for participants to complete. However, our annotations only cover participants in the TGC Classic race, where runners must complete a 128-kilometer course in under 30 hours. The original dataset includes annotations for almost 600 participants at six RPs. Our experiments focused on the final three RPs, denoted as RP1, RP2, and RP3, respectively. These RPs yielded data from beyond kilometer 84, enabling us to assess how well our models performed in the latter part of the race.

Consequently, our study concentrated on two significant race stages, namely $RP1 \rightarrow RP2$ and $RP2 \rightarrow RP3$. \todo{These stages represent the first and second RPs and the second and third RPs, respectively, and are crucial for understanding the runners' fatigue and their ability to improve their running style under the pressure of visualizing the finish line at RP3}. The stage $RP1 \rightarrow RP2$ covers 25km and the stage $RP2 \rightarrow RP3$ covers 15km. Due to the variability in participants' performance, the gap between the first and last runners widened along the course, and the number of participants decreased. Out of the initial dataset of almost 600 participants, we selected 214 participants who appeared in both $RP1 \rightarrow RP2$ and $RP2 \rightarrow RP3$. Among them, 129 participants covered the $RP1 \rightarrow RP2$ stage, and 111 covered the $RP2 \rightarrow RP3$ stage. Participants who only appeared at one RP for each stage were considered negative samples during training. For each participant, we fed seven-second clips at 25 frames per second from each recording point into the footage pre-processing block, as described in Section~\ref{sec:backbones}, using the same frames per second recommended by Carreira and Zisserman~\cite{Carreira17}.

\subsection{Experimental Setup}

This section describes the evaluation protocol used for re-ID tasks in our study. We aim to evaluate the performance of various re-ID models on a large-scale dataset. To this end, we used two evaluation metrics.

Mean Average Precision (mAP) is a widely used evaluation metric for re-ID tasks. It measures the average precision across all possible rankings of the images. mAP is calculated by computing the average precision (AP) for each class and then taking the mean of these APs over all classes. AP is defined as the area under the precision-recall curve (PR curve) for a given query image.

The CMC curve is another evaluation metric used in re-ID tasks. It measures the percentage of correct matches at each retrieved image rank. The CMC curve is obtained by computing the percentage of correct matches for each rank and plotting the results on a graph. A robust re-ID model should have a high CMC curve, indicating that the correct match will likely be found among the top-ranked images.

To ensure the validity of our results, we employed \mbox{10-fold} cross-validation. This approach divides the dataset into ten folds, each containing an equal number of samples. In each fold, one subset of the data is used as the test set, and the remaining nine subsets are used as the training set. This process is repeated ten times, each subset serving as the test set once. The performance metrics (mAP and CMC curve) are averaged across all the folds to obtain the final evaluation scores.

The 10-fold cross-validation scheme, while being constrained due to the limited number of test samples per fold caused by the nature of the dataset, still proves to be a valuable evaluation protocol for our re-ID model. With race stages containing 111 and 129 positive pairs, only 11 to 12 test samples are available per fold for each stage. However, the 10-fold cross-validation approach enables us to evaluate the model's performance on diverse subsets of the dataset, ensuring that the evaluation results are unbiased towards any subset. The mAP and CMC curves provide crucial insights into the model's strengths and weaknesses.

In summary, we used mAP and CMC curves as evaluation metrics for re-ID tasks and employed 10-fold cross-validation to ensure the validity of our results. These metrics provide a comprehensive evaluation of the performance of the re-ID models on a large-scale dataset.

\begin{figure}[h!]
\centering
\begin{subfigure}[b]{0.5\textwidth}
    \centering
   \includegraphics[scale=0.5]{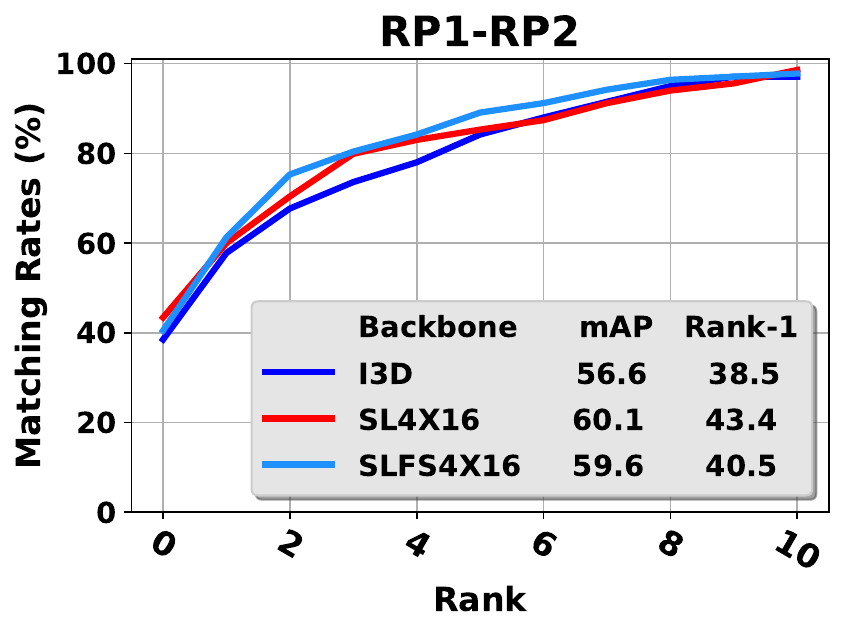}
   \caption{Triplets - CMC considering RP1 as probe and RP2 as gallery.}
   \label{fig:3nnrp1rp2} 
\end{subfigure}

\begin{subfigure}[b]{0.5\textwidth}
    \centering
   \includegraphics[scale=0.5]{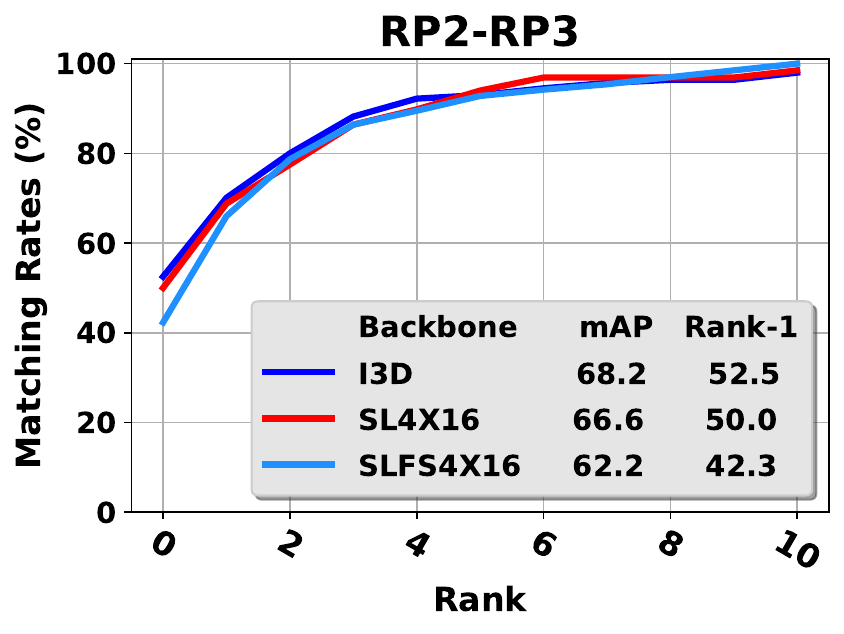}
   \caption{Triplets - CMC considering RP2 as probe and RP3 as gallery.}
   \label{fig:3nnrp2rp3} 
\end{subfigure}

\begin{subfigure}[b]{0.5\textwidth}
    \centering
   \includegraphics[scale=0.5]{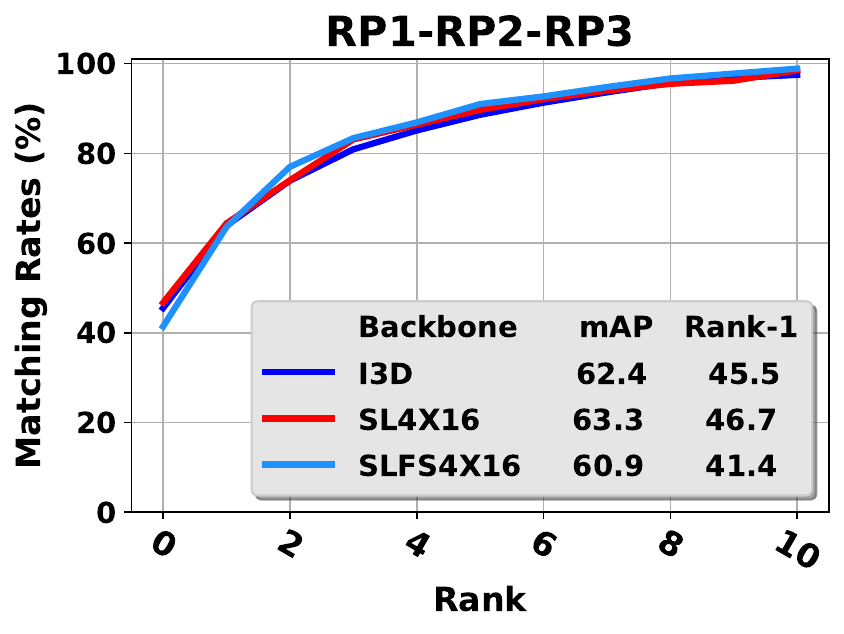}
   \caption{Triplets - Average CMC.}
   \label{fig:3nnrp1rp2rp3} 
\end{subfigure}

\caption{\textbf{The CMC curves of the best-performing models are presented for different RP configurations when triplet loss is used}. SL4X16 and SLFS4X16 correspond to Slow4X16 and SlowFast4x16, respectively, as listed in Table~\ref{table1}.}
\label{fig:cmc3nn}
\end{figure}

\begin{figure}[h!]
\centering
\begin{subfigure}[b]{0.5\textwidth}
    \centering
   \includegraphics[scale=0.5]{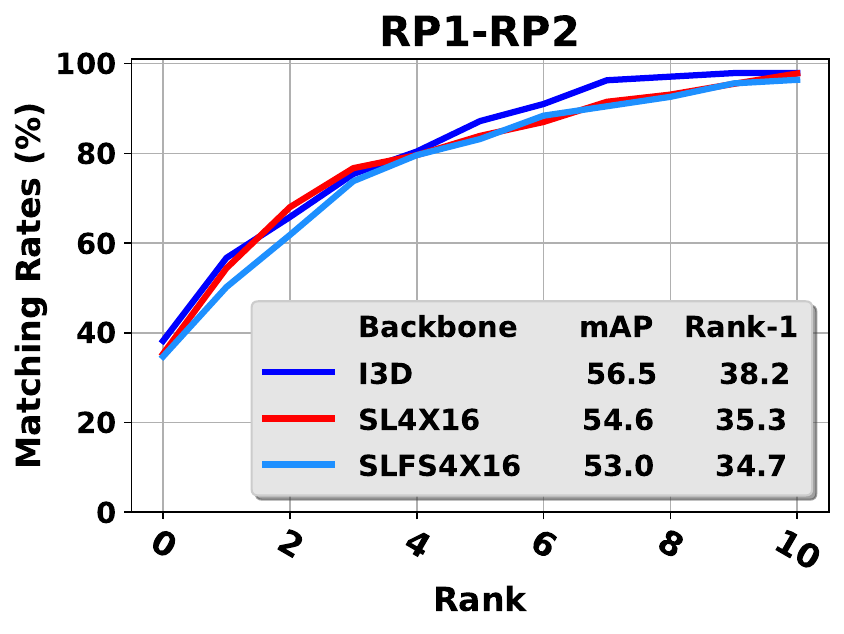}
   \caption{Quadruplets - CMC considering RP1 as probe and RP2 as gallery.}
   \label{fig:4nnrp1rp2} 
\end{subfigure}

\begin{subfigure}[b]{0.5\textwidth}
    \centering
   \includegraphics[scale=0.5]{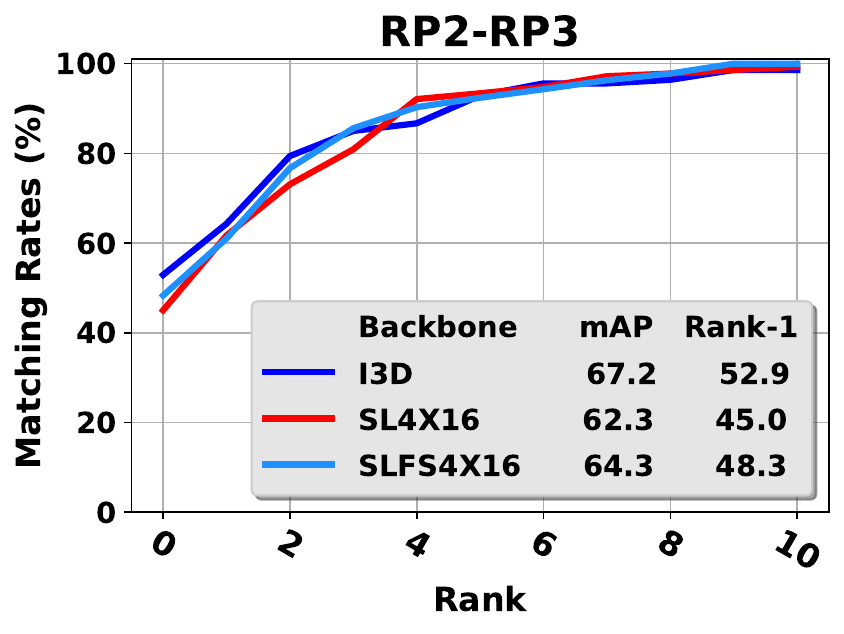}
   \caption{Quadruplets - CMC considering RP2 as probe and RP3 as gallery.}
   \label{fig:4nnrp2rp3} 
\end{subfigure}

\begin{subfigure}[b]{0.5\textwidth}
    \centering
   \includegraphics[scale=0.5]{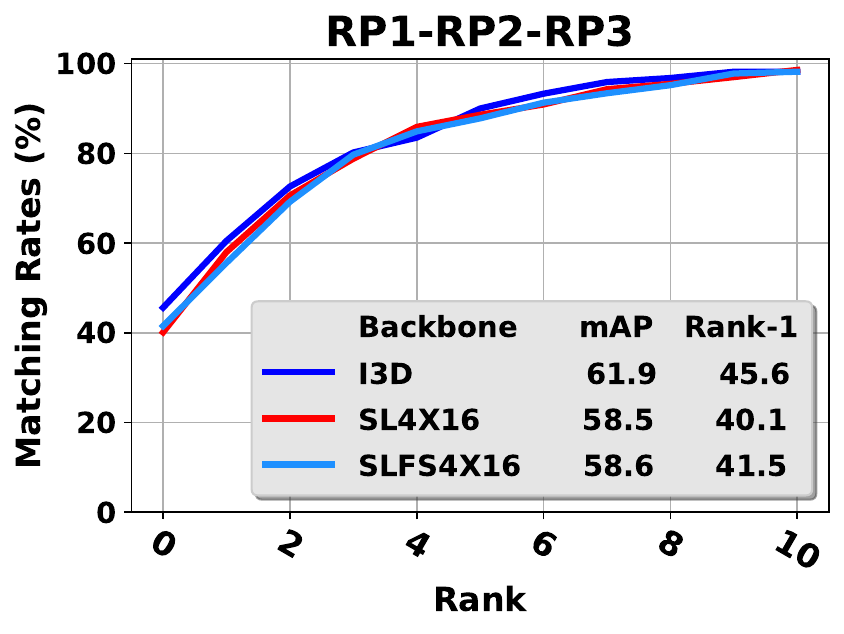}
   \caption{Quadruplets - Average CMC.}
   \label{fig:4nnrp1rp2rp3} 
\end{subfigure}

\caption{\textbf{The CMC curves of the best-performing models are presented for different RP configurations when quadruplet loss is used}. SL4X16 and SLFS4X16 correspond to Slow4X16 and SlowFast4x16, respectively, as listed in Table~\ref{table1}.}
\label{fig:cmc4nn}
\end{figure}

\subsection{Experimental Evaluation}
Table~\ref{table1} presents the performance of several backbone models on two different loss functions, namely triplet loss and quadruplet loss, as measured by mAP. The table consists of 12 rows of the HAR models (see Section~\ref{sec:backbones}), where each row provides the mAP values for both triplet and quadruplet losses. The first column denotes the backbone models, and the second column shows the number of frames used by the model to generate the gait embeddings (see Section~\ref{sec:preprocessing}). The subsequent columns display the mAP values for each model and loss function for two different race stages, $RP1 \rightarrow RP2$ and $RP2 \rightarrow RP3$, as well as the average mAP across these reference stages.

The results show that Slow 4x16 is the best model for triplets, achieving the highest mAP value of 63.3\% across both reference points, a significant improvement compared to other models. On the other hand, I3D is the best model for quadruplets, with a mean mAP of 61.9\%, which is considerably better than other models. Specifically, I3D outperforms all other models, including I3D NLN, an I3D variant designed to address the covariance shift problem in the batch normalization layer. These results suggest that Slow 4x16 and I3D are well-suited for triplet and quadruplet loss functions and can perform better than other models.

The presented table reveals that the number of frames incorporated in the HAR model does not significantly affect the model's performance. This is evidenced by the SlowFast 8x8 model, which uses more frames and performs worse than the SlowFast 4x16 model for both loss functions. On the other hand, models with fewer frames, such as X3D XS and X3D S, exhibit lower performance compared to models with more frames, including X3D M and X3D L. Thus, it appears that the model's architecture, rather than the amount of input information, is more relevant to performance outcomes.

Finally, the last row of the table presents the average mAP for each model and loss function across both reference stages. The mAP values of the athlete re-ID system indicate that the system performed better in the second stage (RP2 to RP3) compared to the first stage (RP1 to RP2) for both triplet and quadruplet loss. When considering triplet loss, the mAP average was 53.2\% for the first stage and 59.6\% for the second stage. Similarly, the mAP average was 50.5\% for the first stage and 59.3\% for the second stage when considering quadruplet loss. These results suggest that the system can identify athletes with better accuracy as they progress through the race. 
\todo{Studies have previously demonstrated that fatigue can significantly impact a runner's gait, resulting in a hunched posture, shorter strides, and decreased speed~\cite{Bailey2018UnderstandingTI,Riazati2020FatigueIC}. Additionally, fatigue can compromise a runner's coordination and balance, leading to an increased risk of falls or stumbles. However, these studies were conducted under controlled conditions with fresh subjects or limited duration. In contrast, our study examines runners who have already completed 84 kilometers in uncontrolled competition conditions. While fatigue may contribute to changes in gait, runners may also alter their foot strike pattern and increase their cadence as they near the finish line to maintain speed and finish strong, which could impact re-ID accuracy.}
It is also worth noting that all models agree that the second stage provides better accuracy than the first. This consistency across different models suggests that the improved accuracy in the second stage is not simply due to chance or noise in the data. \todo{Instead, it provides further evidence that reaching the CP plays a role in the accuracy of the HAR-based re-ID system.}

The consistency between the mAP values and the CMC curves for triplets and quadruplets is notable in Figures~\ref{fig:cmc3nn} and~\ref{fig:cmc4nn}. The rank-1 accuracy in both curves for the top-performing models is higher in the second stage than in the first stage, corroborating the \todo{CP} effect on athlete re-ID. For instance, the best-performing models in terms of mAP, which uses the triplet loss, achieve a rank-1 accuracy of 52.5\% in the second stage (see Figure~\ref{fig:3nnrp2rp3}), while the rank-1 accuracy for the first stage is 43.4\% (see Figure~\ref{fig:3nnrp1rp2}). Similarly, the best-performing models for the quadruplet loss shows an increase in rank-1 accuracy from 38.2\% in the first stage to 52.9\% in the second stage (see Figures~\ref{fig:4nnrp1rp2} and~\ref{fig:4nnrp2rp3}). These results indicate that the performance of the models is consistent with the expected effect of \todo{CP} on athlete re-ID. The CMC curves also reveal that the models based on the quadruplet loss have lower accuracy than those based on the triplet loss. This finding is consistent with the mAP results in Table~\ref{table1}, which showed a lower performance for the models using the quadruplet loss. Nonetheless, the accuracy of the models based on the quadruplet loss increases in the second stage, indicating that the \todo{CP} effect is also present in these models. Overall, the consistency between the mAP values and the CMC curves demonstrates the reliability of the experimental setup and confirms the impact of athlete \todo{CP} on re-ID performance.

\section{Conclusion}

This study explored HAR models in athlete re-ID and showed promising results. Using HAR models allows for extracting robust features from pose sequences and can handle noisy skeleton data in realistic scenes. The triplet and quadruplet loss functions were evaluated, and it was found that the second stage of the race provides better accuracy than the first stage, likely due to athletes reaching the CP. The consistency between the mAP and CMC curves across the three top HAR models highlights the reliability of these findings.

Furthermore, the study provides insight into choosing an appropriate loss function when using HAR for athlete re-ID. While both the triplet and quadruplet loss functions were practical, the triplet loss function provided slightly better results overall. However, it should be noted that using a more complex loss function requires more computational resources, so the choice of loss function should be balanced with practical considerations. Overall, the findings of this study inform the development of more robust and accurate athlete re-ID systems, with potential applications in sports events and security settings.

{\small
\bibliographystyle{ieee}

\begin{thebibliography}{10}\itemsep=-1pt

\bibitem{Ahmed20}
F.~Ahmed, A.~S. M.~H. Bari, and M.~L. Gavrilova.
\newblock Emotion recognition from body movement.
\newblock {\em IEEE Access}, 8:11761--11781, 2020.

\bibitem{Bailey2018UnderstandingTI}
J.~P. Bailey, J.~S. Dufek, J.~F. Silvernail, J.~W. Navalta, and J.~Mercer.
\newblock Understanding the influence of perceived fatigue on coordination
  during endurance running.
\newblock {\em Sports Biomechanics}, 19:618 -- 632, 2018.

\bibitem{Ben-Ami12-bmvc}
I.~Ben-Ami, T.~Dekel, and S.~Avidan.
\newblock Racing bib number recognition.
\newblock In {\em British Machine Vision Conference}, pages 1--10, Surrey, UK,
  2012. British Machine Vision Association.

\bibitem{Carissimi18}
N.~Carissimi, P.~Rota, C.~Beyan, and V.~Murino.
\newblock Filling the gaps: Predicting missing joints of human poses using
  denoising autoencoders.
\newblock In {\em ECCV Workshops}, 2018.

\bibitem{Carreira17}
J.~Carreira and A.~Zisserman.
\newblock {Quo Vadis, Action Recognition? A New Model and the Kinetics
  Dataset}.
\newblock In {\em 2017 IEEE Conference on Computer Vision and Pattern
  Recognition (CVPR)}, pages 4724--4733, 2017.

\bibitem{Chen2017}
W.~Chen, X.~Chen, J.~Zhang, and K.~Huang.
\newblock {Beyond triplet loss: a deep quadruplet network for person
  re-identification}, 2017.

\bibitem{Yapkan21}
Y.~Choi, Y.~Napolean, and J.~C. van Gemert.
\newblock The arm-swing is discriminative in video gait recognition for athlete
  re-identification.
\newblock In {\em 2021 IEEE International Conference on Image Processing
  (ICIP)}, pages 2309--2313, 2021.

\bibitem{Chokchai18}
C.~Chokchai.
\newblock Low cost and high performance uhf rfid system using arduino based on
  iot applications for marathon competition.
\newblock In {\em 2018 21st International Symposium on Wireless Personal
  Multimedia Communications (WPMC)}, pages 15--20, 2018.

\bibitem{Cioppa22}
A.~Cioppa, S.~Giancola, A.~Deli{\`e}ge, L.~Kang, X.~Zhou, Z.~Cheng, B.~Ghanem,
  and M.~V. Droogenbroeck.
\newblock {SoccerNet-Tracking: Multiple Object Tracking Dataset and Benchmark
  in Soccer Videos}.
\newblock {\em 2022 IEEE/CVF Conference on Computer Vision and Pattern
  Recognition Workshops (CVPRW)}, pages 3490--3501, 2022.

\bibitem{Enck10}
W.~Enck, P.~Gilbert, B.-G. Chun, L.~P. Cox, J.~Jung, P.~Mcdaniel, and A.~Sheth.
\newblock Taintdroid: An information-flow tracking system for realtime privacy
  monitoring on smartphones.
\newblock {\em ACM Trans. Comput. Syst.}, 32:5:1--5:29, 2010.

\bibitem{Fan20}
C.~Fan, Y.~Peng, C.~Cao, X.~Liu, S.~Hou, J.~Chi, Y.~Huang, Q.~Li, and Z.~He.
\newblock Gaitpart: Temporal part-based model for gait recognition.
\newblock In {\em 2020 IEEE/CVF Conference on Computer Vision and Pattern
  Recognition (CVPR)}, pages 14213--14221, 2020.

\bibitem{X3D20}
C.~Feichtenhofer.
\newblock X3d: Expanding architectures for efficient video recognition.
\newblock {\em 2020 IEEE/CVF Conference on Computer Vision and Pattern
  Recognition (CVPR)}, pages 200--210, 2020.

\bibitem{SlowFast19}
C.~Feichtenhofer, H.~Fan, J.~Malik, and K.~He.
\newblock Slowfast networks for video recognition.
\newblock {\em 2019 IEEE/CVF International Conference on Computer Vision
  (ICCV)}, pages 6201--6210, 2018.

\bibitem{Slow21}
C.~Feichtenhofer, H.~Fan, B.~Xiong, R.~B. Girshick, and K.~He.
\newblock A large-scale study on unsupervised spatiotemporal representation
  learning.
\newblock {\em 2021 IEEE/CVF Conference on Computer Vision and Pattern
  Recognition (CVPR)}, pages 3298--3308, 2021.

\bibitem{freire22icpr}
D.~Freire-Obreg{\'o}n, J.~Lorenzo-Navarro, O.~J. Santana,
  D.~Hern{\'a}ndez-Sosa, and M.~Castrill{\'o}n-Santana.
\newblock {Towards cumulative race time regression in sports: I3D ConvNet
  transfer learning in ultra-distance running events}.
\newblock {\em 2022 26th International Conference on Pattern Recognition
  (ICPR)}, pages 805--811, 2022.

\bibitem{Freire20}
D.~Freire-Obregón, M.~Castrillón-Santana, P.~Barra, C.~Bisogni, and M.~Nappi.
\newblock An attention recurrent model for human cooperation detection.
\newblock {\em Computer Vision and Image Understanding}, 197-198:102991, 2020.

\bibitem{Galbally14}
J.~Galbally, S.~Marcel, and J.~Fierrez.
\newblock Image quality assessment for fake biometric detection: Application to
  iris, fingerprint, and face recognition.
\newblock {\em IEEE Transactions on Image Processing}, 23:710--724, 2014.

\bibitem{Goodrich21}
M.~T. Goodrich, S.~Gupta, H.~Khodabandeh, and P.~Matias.
\newblock How to catch marathon cheaters: New approximation algorithms for
  tracking paths, 2021.

\bibitem{Carrascosa20}
P.~Hernández-Carrascosa, A.~Penate-Sanchez, J.~Lorenzo-Navarro,
  D.~Freire-Obregón, and M.~Castrillón-Santana.
\newblock {TGCRBNW: A Dataset for Runner Bib Number Detection (and Recognition)
  in the Wild}.
\newblock In {\em 2020 25th International Conference on Pattern Recognition
  (ICPR)}, pages 9445--9451, 2021.

\bibitem{Hossain12}
E.~Hossain, G.~Chetty, and R.~G{\"o}cke.
\newblock {Multi-view Multi-modal Gait Based Human Identity Recognition from
  Surveillance Videos}.
\newblock In {\em IAPR TC3 Workshop on Multimodal Pattern Recognition of Social
  Signals}, 2012.

\bibitem{Lingjia19}
L.~Huang, H.~Ma, W.~Yan, W.~Liu, H.~Liu, and Z.~Yang.
\newblock Sports motion recognition based on foot trajectory state sequence
  mapping.
\newblock In {\em 2019 International Joint Conference on Neural Networks
  (IJCNN)}, pages 1--8, 2019.

\bibitem{Jain04}
A.~Jain, A.~Ross, and S.~Prabhakar.
\newblock An introduction to biometric recognition.
\newblock {\em IEEE Transactions on Circuits and Systems for Video Technology},
  14(1):4--20, 2004.

\bibitem{Jiang16}
H.~Jiang and E.~G. Learned-Miller.
\newblock Face detection with the faster r-cnn.
\newblock {\em 2017 12th IEEE International Conference on Automatic Face \&
  Gesture Recognition (FG 2017)}, pages 650--657, 2016.

\bibitem{Kay17}
W.~Kay, J.~Carreira, K.~Simonyan, B.~Zhang, C.~Hillier, S.~Vijayanarasimhan,
  F.~Viola, T.~Green, T.~Back, P.~Natsev, M.~Suleyman, and A.~Zisserman.
\newblock {The Kinetics Human Action Video Dataset}.
\newblock {\em CoRR}, 2017.

\bibitem{Klontz13}
J.~C. Klontz and A.~K. Jain.
\newblock A case study of automated face recognition: The boston marathon
  bombings suspects.
\newblock {\em Computer}, 46(11):91--94, 2013.

\bibitem{Lei19}
Q.~Lei, J.-X. Du, H.-B. Zhang, S.~Ye, and D.-S. Chen.
\newblock A survey of vision-based human action evaluation methods.
\newblock {\em Sensors}, 19(19), 2019.

\bibitem{Manafifard17}
M.~Manafifard, H.~Ebadi, and H.~A. Moghaddam.
\newblock A survey on player tracking in soccer videos.
\newblock {\em Comput. Vis. Image Underst.}, 159:19--46, 2017.

\bibitem{Nambiar19}
A.~M. Nambiar, A.~Bernardino, and J.~C. Nascimento.
\newblock {Gait-based Person Re-identification}.
\newblock {\em ACM Computing Surveys (CSUR)}, 52:1 -- 34, 2019.

\bibitem{penate20prl}
A.~Penate-Sanchez, D.~Freire-Obregón, A.~Lorenzo-Melián, J.~Lorenzo-Navarro,
  and M.~Castrillón-Santana.
\newblock {TGC20ReId: A dataset for sport event re-identification in the wild}.
\newblock {\em Pattern Recognition Letters}, 138:355--361, 2020.

\bibitem{Riazati2020FatigueIC}
S.~Riazati, N.~Caplan, M.~Matabuena, and P.~R. Hayes.
\newblock Fatigue induced changes in muscle strength and gait following two
  different intensity, energy expenditure matched runs.
\newblock {\em Frontiers in Bioengineering and Biotechnology}, 8, 2020.

\bibitem{Schroff15}
F.~Schroff, D.~Kalenichenko, and J.~Philbin.
\newblock {FaceNet: A unified embedding for face recognition and clustering}.
\newblock In {\em 2015 IEEE Conf. on Computer Vision and Pattern Recognition
  (CVPR)}, pages 815--823, 2015.

\bibitem{C2D14}
K.~Simonyan and A.~Zisserman.
\newblock Two-stream convolutional networks for action recognition in videos.
\newblock {\em ArXiv}, abs/1406.2199, 2014.

\bibitem{Liu19}
K.~Sun, B.~Xiao, D.~Liu, and J.~Wang.
\newblock Deep high-resolution representation learning for human pose
  estimation.
\newblock {\em 2019 IEEE/CVF Conference on Computer Vision and Pattern
  Recognition (CVPR)}, pages 5686--5696, 2019.

\bibitem{Teepe21}
T.~Teepe, A.~Khan, J.~Gilg, F.~Herzog, S.~Hörmann, and G.~Rigoll.
\newblock Gaitgraph: Graph convolutional network for skeleton-based gait
  recognition.
\newblock In {\em 2021 IEEE International Conference on Image Processing
  (ICIP)}, pages 2314--2318, 2021.

\bibitem{Torres22}
L.~Torres-Ronda, E.~Beanland, S.~Whitehead, A.~J. Sweeting, and J.~Clubb.
\newblock {Tracking Systems in Team Sports: A Narrative Review of Applications
  of the Data and Sport Specific Analysis}.
\newblock {\em Sports Medicine - Open}, 8, 2022.

\bibitem{Zakrzewski23}
J.~Tozer and N.~Anderson.
\newblock How can someone who cheated cross a finish line and collect a medal?
\newblock
  \url{https://www.dailymail.co.uk/news/article-11987195/Leading-British-ultra-marathon-runner-47-disqualified-race-using-CAR.html}.
\newblock Accessed: 2023-05-05.

\bibitem{Wang03}
L.~Wang, T.~Tan, H.~Ning, and W.~Hu.
\newblock Silhouette analysis-based gait recognition for human identification.
\newblock {\em IEEE Trans. Pattern Anal. Mach. Intell.}, 25:1505--1518, 2003.

\bibitem{NonlocalNN17}
X.~Wang, R.~B. Girshick, A.~K. Gupta, and K.~He.
\newblock Non-local neural networks.
\newblock {\em 2018 IEEE/CVF Conference on Computer Vision and Pattern
  Recognition}, pages 7794--7803, 2017.

\bibitem{Wnuk10}
K.~Wnuk and S.~Soatto.
\newblock Analyzing diving: {A} dataset for judging action quality.
\newblock In R.~Koch and F.~Huang, editors, {\em Computer Vision - {ACCV} 2010
  Workshops - {ACCV}}, volume 6468 of {\em Lecture Notes in Computer Science},
  pages 266--276. Springer, 2010.

\bibitem{zhang2021bytetrack}
Y.~Zhang, P.~Sun, Y.~Jiang, D.~Yu, Z.~Yuan, P.~Luo, W.~Liu, and X.~Wang.
\newblock {ByteTrack: Multi-Object Tracking by Associating Every Detection
  Box}.
\newblock In {\em European Conference on Computer Vision}, 2021.

\end{thebibliography}

}

\end{document}